# Investigation of Color Constancy for Ubiquitous Wireless LAN/Camera Positioning: An Initial Outcome

[1]Wan Mohd Yaakob Wan Bejuri, [2]Mohd Murtadha Mohamad, [3]Maimunah Sapri, [4]Mohd Adly Rosly

[1, 2,4]*Faculty of Computer Science & Information Systems, Universiti Teknologi Malaysia*
[3]*Centre of Real Estate Studies, Universiti Teknologi Malaysia*
[1]*wanmohdyaakob@ieee.org,* [2]*murtadha@utm.my,* [3]*maimunahsapri@utm.my,*
[4]*madlysalaf@gmail.com*

## Abstract

   *This paper present our color constancy investigation in the hybridization of Wireless LAN and Camera positioning in the mobile phone. Five typical color constancy schemes are analyzed in different location environment. The results can be used to combine with RF signals from Wireless LAN positioning by using model fitting approach in order to establish absolute positioning output. There is no conventional searching algorithm required, thus it is expected to reduce the complexity of computation. Finally we present our preliminary results to illustrate the indoor positioning algorithm performance evaluation for an indoor environment set-up.*

## 1. Introduction

   The movement into the ubiquitous computing realm will integrate the advances from both mobile and pervasive computing. Though these terms are often used interchangeably, they are conceptually different and employ different ideas of organizing and managing computing services [1] [2] [3]. The increasing of mobile usage has been widespread since it is a most cool gadget that compromise mobility, efficiency and effective to the end users. Nowadays, mobile phone can be considered as a multi-purpose device. One of the function of mobile device, it can provide navigation services to end user. The navigation system  in the mobile device rely mainly on GPS for absolute position determination. However, the positioning using GPS on mobile can get a 10+-m accuracy in the best environment. This accuracy becomes worse in urban areas especially inside building [4] [5] [6] [7] [8] [9] [10] [11] [12] [13]. The navigation in that particular area becomes a more challenging task as the user moves in general in areas where no one of the common location techniques works continuously in standalone mode. In order to solve this challenging task of continuous position determination, there is a need of combination with other location techniques.

   Although the integration of GPS with other external positioning technology can be made, this solution is not intended very much to solve this issue since it may make end users feel harsh on device integration. On the other side, others sensor on mobile phone such as WLAN, Camera and Bluetooth can be used as alternative positioning method. The integration of positioning within mobile phone can be made to ensure mobile navigation technologies become more pervasive, mobility and ubiquity. In this paper, we focus on the indoor positioning in term to how to get positioning information inside building in a hallways/corridor by using hybridization of WLAN and Camera. We do not cover positioning outside building and indoor/outdoor positioning since the indoor positioning is the crucial part in the navigation system. As addition, the integration of GPS with alternative positioning can be made easily. The structure of the paper is as follows. Section 2 will present the reviews of color constancy. Section 3 will present our proposed method. The details of our preliminary result are covered in Section 4. Finally, discussion and research direction are given in Section 5.

## 2. Review of color constancy

The measurement of color constancy can be estimated by using the light source color. In this estimation, the original image values are transformed in order to correct the chromaticity of the light source. Let say image values $f = (f_R, f_G, f_B)^T$ , this value depends on light source color ($\lambda$), the function of camera sensitivity (see equation 1) and reflectance of surface $(x, \lambda)$ , where $\lambda$ equals to light wavelength and x is the coordinate space [14] [15] (see equation 2);

$$P(\lambda) = \left( P_R(\lambda), P_G(\lambda), P_B(\lambda) \right)^T \tag{1}$$

$$f_c = m_b(x) \int_w I(\lambda) p_c(\lambda) S(x,\lambda) d\lambda + m_s(x) \int_w I(\lambda) p_c d\lambda \tag{2}$$

which are; visible spectrum $c = \{R, G, B\}$, factor of scale $m_b$ and $m_s$ that model the body relative amount and reflectance of specular which contribute overall reflectance of light (which located at x). Under the Lambertian assumption the specular reflection is ignored. This result in the following model (see equation 3), which are $m(x)$ is the shading value of Lambertian . Assumed that the scene is illuminated by a single of light source and the observed color of the light source e, which it depends on the light source of color $I(\lambda)$, and the sensitivity of camera function $p(x)$:

$$f_c(x) = m(x) \int_w I(\lambda) p_c(\lambda) S(x,\lambda) d\lambda \tag{3}$$

$$e = \begin{pmatrix} e_R \\ e_G \\ e_B \end{pmatrix} = \int_w I(\lambda) p(\lambda) d(\lambda) \tag{4}$$

If the value in both of $I(\lambda)$ and $p(\lambda)$ is unknown, and therefore the estimate of value e is an under constrained problem that can not be solved without further assumptions. In this part, we will explain briefly on how to correct image in which is very crucial to estimate the light source color. In order to do that, the diagonal change or von Kries model [16] is used, which is operated by not changing the basic color [17] [18] or using a sharpening of the spectrum [19] [20]. These techniques have shown their merits to improve the quality of output images, if the source of light in which the original image was recorded is known. Diagonal model used is given by equation (5), where $f_u$ is the image taken in unknown light source is, $f_t$ is the same image to change which that appears if it has been taken under canonical illuminant environment, and $D_{u,t}$ is a diagonal matrix that maps the colors that brought under one light source is not known u for their corresponding colors under statutory light source $c$ (see equation 6).

$$f_t = D_{u,t} f_u \tag{5}$$

$$\begin{pmatrix} R_c \\ G_c \\ B_c \end{pmatrix} = \begin{pmatrix} d_1 & 0 & 0 \\ 0 & d_2 & 0 \\ 0 & 0 & d_3 \end{pmatrix} \begin{pmatrix} R_u \\ G_u \\ B_u \end{pmatrix} \tag{6}$$

Although this model is merely an estimate of the illumination of light source, and may not accurately able to demonstrate the photometric changes. However, it may also be accepted as color correction model [21] [22] [23] [24] [25] and it enhances a lot of color constancy algorithms. Diagonal mapping used throughout this paper creates the output images after correction of the color constancy algorithm, where a perfect white light, which($\frac{1}{\sqrt{3}}, \frac{1}{\sqrt{3}}, \frac{1}{\sqrt{3}})^T$, used as a source statutory light. In this review we only focus on static rules in which it can be categorized low-level based and physics-based.

## 2.1. Low-level based

The most popular method in low level based is Grey world [26]. This type uses assumption in deviation from the color chromaticity scene moderated by the effects of light sources (or illumination effect). The light source color e can be determined by calculate the average color value in a image as shown below:

$$\int f_c(x)dx = ke_c \tag{7}$$

where k is a multiplicative constant such color light source (color of illuminant), $e = (e_R, e_G, e_B)^T$, has a length unit. In other side, assumption by using white patch is another popular one [27] which executed based on RGB channels response (maximum value) in which is obtained by a perfect reflectance on the surface. In practice, the assumption of perfect reflectance value is reduced by considering the separate color channels, resulting in the max-RGB algorithm. This method is to determine the illuminant of light source (illuminant value) by calculating the maximum response in separate color channels:

$$\underset{x}{max} f_c(x) = ke_c \tag{8}$$

Previous algorithms were using some kind of smooth technique as a pre process in illumination determination [28] [29]. However, this step is almost similar in term of performance of the White-Patch method as segmentation by using Grey World. As additional, the advantage of the Local Space Average color method [29] can be used to determine illuminant pixel wise in which it does not require an input image that can be obtained in uniform light environment. By referring to [30] [31], we can see the result of max-RGB algorithm which shows the dynamic range of image have an enormous influence in term of performance. The White Patch and Gray World shown as special instantiations of a more general framework (framework of Minkowski) [32]:

$$\mathcal{L}_c(p) = (\int f_c^p(x)dx)^{\frac{1}{p}} = ke_c \tag{9}$$

In equation (9), the replacement of $p = 1$ is equal to calculate and obtain the average of f (x). In general, p is tuned to the data set at hand in order to obtain a suitable variable. Therefore, the optimum variable of this parameter may vary for different data sets. The assumptions of constancy of color are based on the color distributions which appear in the image. According to [33], incorporated of higher image statistics in the image derivatives form, where the Grey-Edge framework set combines well-known methods such as equations on (9), and methods based on first order and second derivatives:

$$\left(\int |\frac{\partial^n f_{c,\sigma}(x)}{\partial x^n}|^p \ dx\right)^{\frac{1}{p}} = ke_c^{n,p,\sigma}, \tag{10}$$

where, $|.|$ means the Frobenius norm, $c = \{R, G, B\}, n$ in order derivatives and Minkowski p-norm. In addition, derivatives are defined as convolving the image by Gaussian filter with parameter of scale (σ) [62] as shown below:

$$\frac{\partial^{s+t} f_{c,\sigma}}{\partial x^s y^t} = f_c * \frac{\partial^{s+t} G_\sigma}{\partial x^s \partial y^t} \tag{11}$$

where, * means convolution and $s + t = n$. This method is updated in [34] by using constraints illuminant. In addition in [35], the model of the spatial dependencies between pixel have been made and it shows that this method can learn dependencies between pixel more effective compared to Grey-Edge method. The extension of the research finally made in [36], in order to establish a physics-based weighting scheme since some of edge may contain a diverse information. The results show this method is more accurate but lacking in complexity cost.

## 2.2. Physic-based method

Currently, most of the methods are based on model Lambertian (please see equation 2), however there are some methods or techniques which are based on dichromatic reflection model (please see equation 1). Basically, this kind of method uses information physical interaction between light source by utilizing dichromatic model to illuminants constraint. If multiple of such planes are found, corresponding to various different surfaces, then the color of the light source is estimated using the intersection of those planes. There are previous proposed approaches that use highlights [37] [38] [39] [40]. Various approaches have been proposed that the use specularities or highlights [37] [38] [39] [40]. Intuition behind these rules is that if the pixels found in the body reflectance factor mb in eq. (1) is near zero (0), then the color of the pixels are similar or the same color light sources. However, all these methods have some shortcomings: take the challenge and the specular reflection color sections may also occur. Effectively eliminate the usability of the last pixel specular (more may be cut from other pixels). Methods based on different physical and is suggested in [41]. This method uses the dichromatic reflection model to project the color of a single pixel in the chromaticity space. Then, a set of light sources may be modeled using a Planckian black-body radiator. Planckian locus intersects the surface of the dichromatic line for the light source color. This method, in theory, allows the estimated illuminant of the surface although there is only one present at the scene. However, it requires all the pixels in the image to be divided, so that all surfaces are uniquely identified. Alternatively, the colors in the image can be described using multilinear models consisting of several simultaneous planes oriented about an axis defined by the illuminant [42] [43]. This eliminates the problem of pre-segmentation, but does not depend on the observation that the color representative of any particular substance can be identified. In [44], this requirement is relaxed, which resulted in two Hough transform voting procedures.

## 3. Methodology

Generally, our concept is very simple which is to find any visual landmarks and match them with information that obtained from WLAN positioning in order to estimate indoor positioning. By doing that, we expect to compute precisely the final positioning output. The available visual landmarks at the hallway are not the same type with open area such as corners, doors and transactions between floors to wall. However, the features have homogeneity high degree, in which it is not uniquely identified but it can be solve by comparing with known feature location. In our scope, we only discuss how to determine user positioning in hallway only. By refer to Figure 1, it shows our algorithm flow diagram. Our flow diagram basically uses two (2) different sensor input which are camera and WLAN receiver. The integration of different input is expected to establish the final positioning output be more accurate rather than positioning by single positioning sensor. The input from camera which is obtained by capturing the image will be extracted. In the image, we will extract visible feature interest point (corner), besides that, we also obtain WLAN positioning coordinate from WLAN signal strength. This type of two (2) different information will be sent directly by using our mobile phone trough existing wireless network to our server. In our server, the image input will firstly processed by color constancy algorithm in order to reduce illumination error in the image. Then it will be processed by image segmentation method before it processed through corner detection process in order to get the feature interest point. After image information processing, this interest point image will be matched with coordinate information that are obtained from WLAN positioning by using model fitting approach.

### 3.1. Color constancy-based feature detection

In color constancy-based feature detection part, our algorithm will locate image features as a first step. However, there are some limitation of location information in the building map, the map suppose to have another information including door and corner. Usually this type of information we can call as microlandmarks which are located in the image where the doorways lines and floor edge corners meet. Although the microlandmarks or feature interest point is suffered by illumination change environment in hallways, we prefer to use color constancy-based feature detection in order to obtain such information interest point on the floor edges. By doing that, it is expected that this kind of solution can reduce illumination error on image, and then it is made much easier for another step to recognize feature interest point. The mean shift [45] method is used in order to segment the image. The image segmentation process basically aim to make categorization of object on a image. After the image has been segmented, the final step in this phase is to detect interest corner at edge intersection of doorways and floor by using cornerity metric [46]. However, we believe that this type of corner detection cannot search all suitable corner, but at least it can find many of them. The false detection of corner will not correspond to the floor plan since there is occlusion of the floor.

## 3.2. Feature matching

The final output of previous phase is the coordinate of feature interest point. When this information is obtained, we will integrate these information together with WLAN positioning information. The WLAN [47] will give positioning information 10+-m accuracy in order to select which possible hallways that user should be located. Additionally, it provides the estimated gross center for regional to be tested; matching system can also search radius estimate based on accuracy and some ideas useful horizontal visibility location system camera. The integration of two different information will finally make correspondence between image captured and floor plan. The reason behind this is to reduce ambiguous cases and search space positioning information in database.

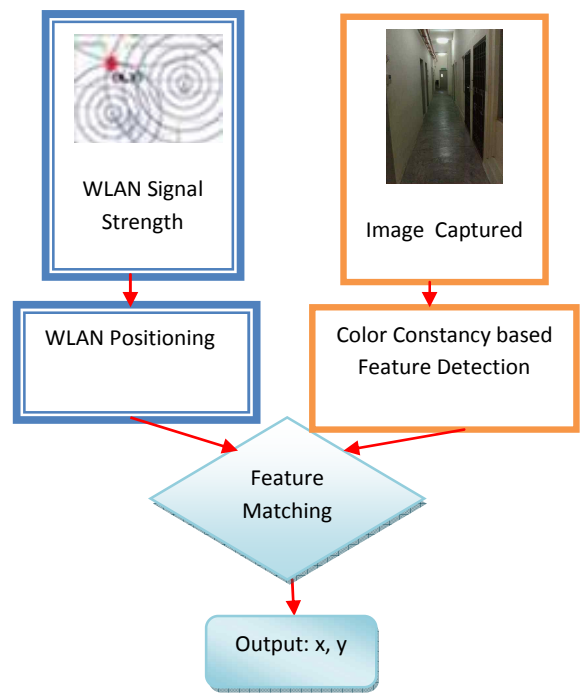

**Figure 1.** Our Flow Diagram to Estimate Positioning on the Hallway by using Two (2) Different Input in order to Generate the Final Positioning Output.

For the correspondence part, there are too much possible correspondence. For example, if we want to match 10 point in a image captured with 32 point in a floor map, the possibble corresponding match between image and floor point may result 4 billion possible four-point correspondences. In order to solve this issue, we prefer to use Random Sample Random Sample Consensus (RANSAC) [48] that operates to select and optimize the hypotheses. In this part, we use minimal structural assumptions to generate the hypotheses for the RANSAC algorithm in order to fit lines in the image-space features, at which produce left and right correspondence line. After that, the algorithm chooses as random two (2) points from each line in image captured that orders the points along the lines consistently. Larger distance between 2 points, means more stable and likely to produce stable camera.

## 4. Preliminary Result

In this section, we show the preliminary result by comparing five (5) types of color constancy algorithm in hybrid WLAN/Camera positioning. In order to carry out a large campaign of measurements on the third (3th) floor of the Faculty of Computer Science and Information Systems, Universiti Teknologi Malaysia, Malaysia. We implemented a processing algoritm which is stated in Section three (3) in a server, meanwhile a mobile phone HTC HD Mini operated as a end user mobile device for capturing an image at selected hallway and transmitter unit of image captured via existing WLAN network. Here, we will analyze our result as follows.

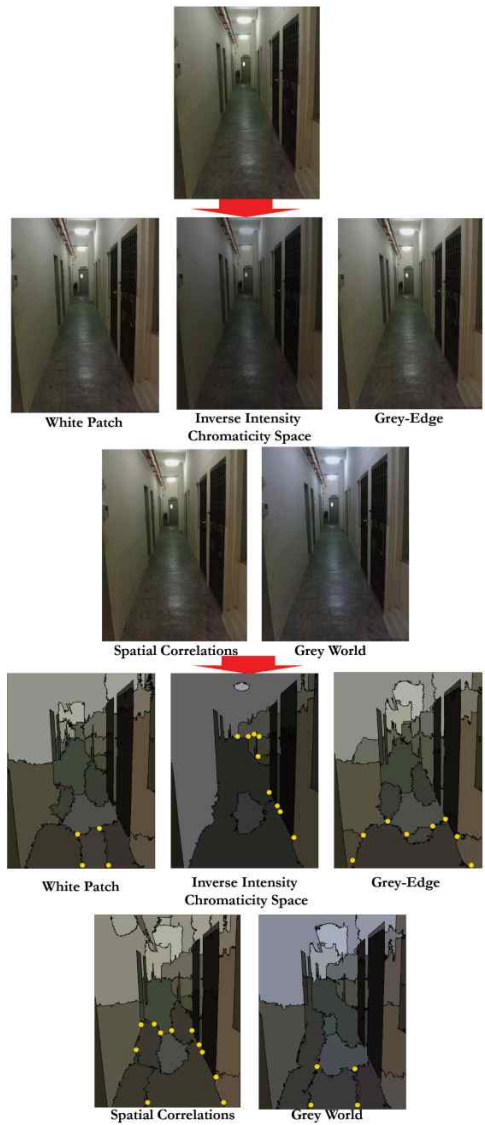

**Figure 2.** Result feature interest point detection at first (1st) location.

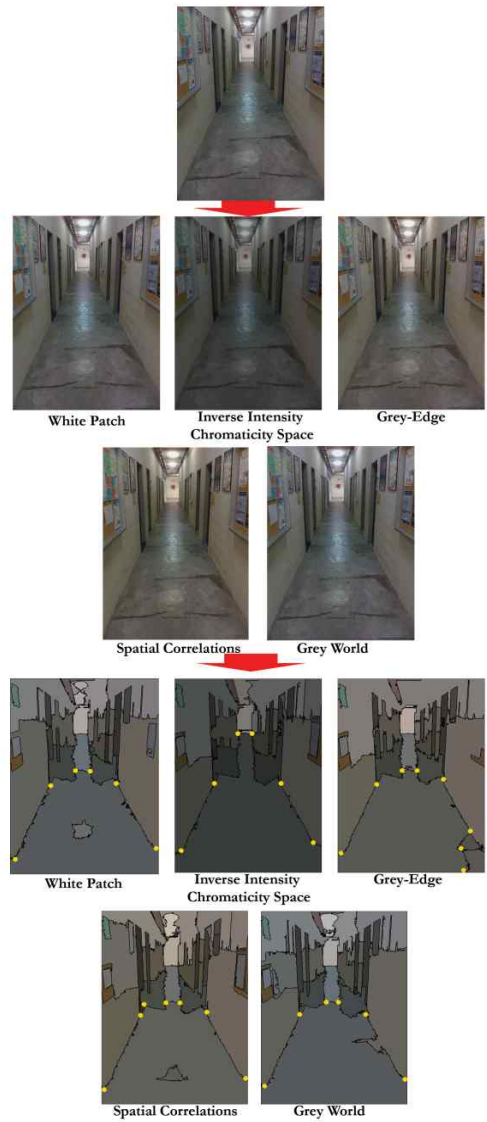

**Figure 3.** Result feature interest point detection at second (2nd) location.

Refer to Figure two (2), the method of five (5) types of feature interest point detection in first (1st) location. The spatial correlation method is the most successful method to locate feature interest point as true detection. However, the methods Inverse Intensity Chromaticity Space just only can detect more at right side and only one (1) at left side. The other method such Grey World, Grey Edge and White Patch can detect interest point on the beginning of floor.

Refer to Figure two (3), the method of five (5) types of feature interest point detection in second (2nd) location. The methods of White Patch, Grey-Edge, Grey World and Spatial Correlations mostly can detect interest feature detection point only a half of floor compared to Inverse Intensity Chromaticity Space method. However, the method of Spatial Correlations can detect more precisely compared to other methods, at intersection of door and floor.

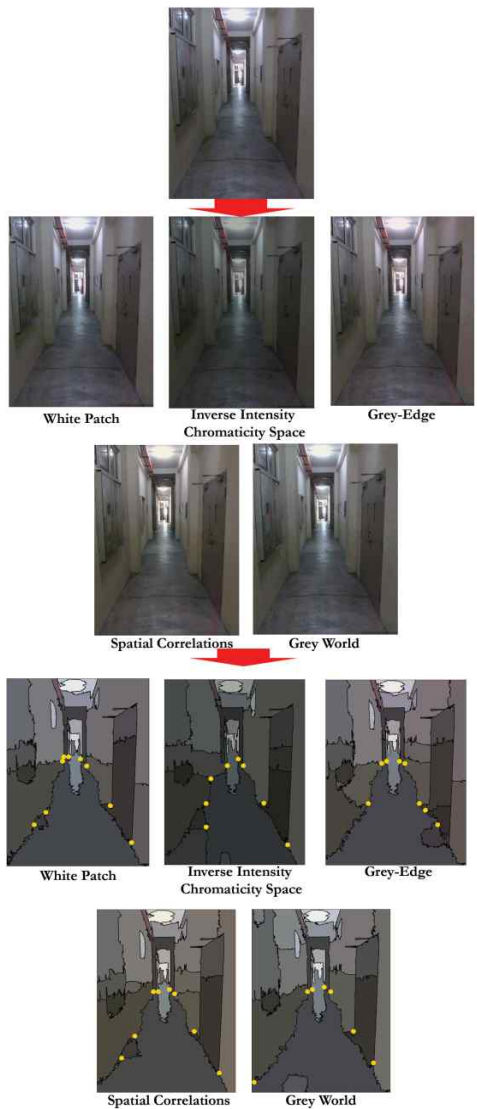

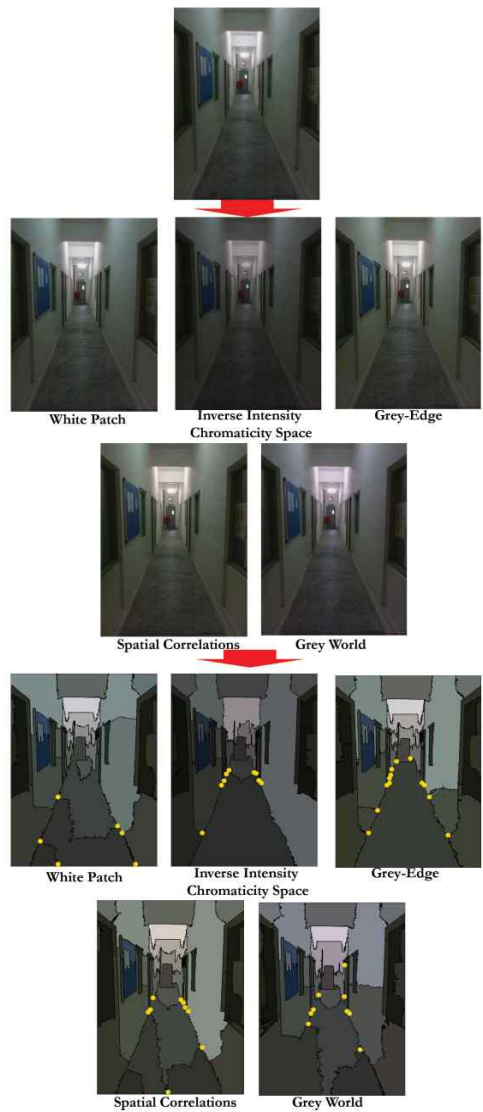

**Figure 4.** Result feature interest point detection at third (3<sup>rd</sup>) location.

**Figure 5.** Result feature interest point detection at forth (4<sup>th</sup>) location.

Refer to Figure four (4), the method of five (5) types of feature interest point detection in third (3th) location. The performance of five (5) methods show the distribution of the interest feature point looks the same. Although the method of Grey World shows it cannot detect accurately on the intersection of floor and door compared to other methods, the method of Inverse Intensity Chromaticity Space looks more worse as overall.

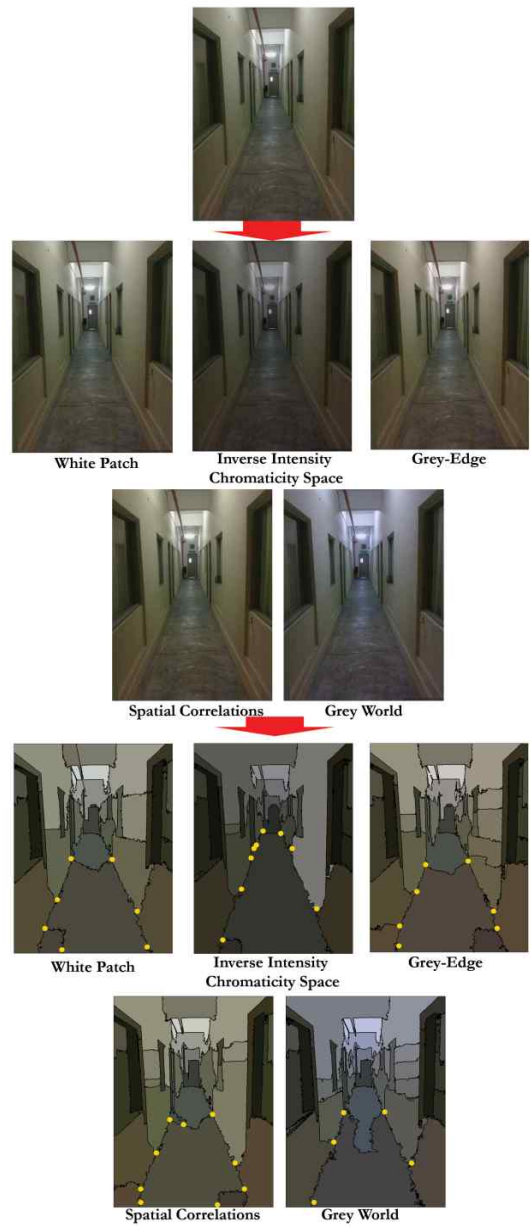

**Figure 6.** Result feature interest point detection at fifth (5$^{th}$) location.

Refer to Figure five (5), the method of five (5) types of feature interest point detection in fourth (4th) location. The method of White Patch is the worse method when applied in this location compared to others. However, the method of Inverse Intensity Chromaticity Space seems to be the best method when applied here and the method of Grey-Edge and Grey World shows its performance in average between White Patch and Spatial Correlations.

Refer to Figure six (6), the method of five (5) types of feature interest point detection in fifth (5th) location. The method of Inverse Intensity Chromaticity Space is the most successful to locate feature interest detection same with previous location. The other methods show the distribution feature interest points almost similar between each other. Although the others methods show the feature detection point can be detected a part of floor only, but it is still better compared to previous location.

## 5. Discussion and future directions

This paper discussed about our preliminary result for using color constancy on hybrid WLAN/Camera positioning. The information from both of sensor output are combined by model fitting approach in order to find the absolute of user target position, without using any conventional searching algorithm. Although we know we did not give quantative result but this result can show how far the color constancy approach can improve our algorithm since it suffered illumination environment. The preliminary result shows, most of the experiment suffered with illumination change. The illumination change will affect position and total of interest points in image. However, it also show the color constancy based on spatial correlation is the most suitable and the most consistence when applied our algorithm. The method of Inverse Intensity Chromaticity Space shows good result in location 2 when the others suffered but it cannot handle in the change environment in which the lightness of image is not consistent.

The hybridization between color constancy maybe applied when end user faces in kind of environment. Although the hybridization between color constancy seems logic, but it is a difficult task for developer as need to think of other criteria on how to switch the color constancy types by classifying which color constancy algorithm is the best fit in a environment. Besides that, developer no need to much worry about the performance of algorithm since the technology of mobile computing is now widely spread with the existence of latest mobile technology such as tablet pc (example: IPAD). However, we will continue our experiment by using this result and simulate with WLAN positioning result in order to get much valuable finding. As a conclusion, interesting future research include investigation of the relation between user localization and hybridization of computational color constancy and WLAN, adopting alternate image correction models used in this paper, and departure of the uniform light source assumption.

## 6. Acknowledgement

This paper was inspired from my Master research project which is related to indoor positioning on mobile phone. The author would also like to thank our supervisor Dr. Mohd Murtadha b Mohamad and also our co-supervisor Dr. Maimunah bt Sapri for their insightful comments on earlier drafts of this paper.

## 7. References

[1]     H. Hamad and F. B. Ibrahim "Path Planning of Mobile Robot Based on Modification of Vector Field Histogram using Neuro-Fuzzy Algorithm", International Journal of Advancements in Computing Technology, Vol. 2, No. 3, 2010.

[2]     H. Wang, "Wireless Sensor Networks for an Extended City Intelligent Transportation System", International Journal of Advancements in Computing Technology, Vol. 3, No.5, 2011.

[3]     N. Sah and N. R. Prakash, "Performance Evaluation of Hand-Off Scheme With Cell Breathing Concept in WCDMA", International Journal of Advancements in Computing Technology, Vol. 1, No. 2, 2009.

[4]     Yanying Gu, A. Lo, and I. Niemegeers, "A Survey of Indoor Positioning Systems for Wireless Personal Networks," IEEE Communications Surveys & Tutorials, vol. 11, no. 1, pp. 13-32, 2009.

[5]     T. Ruiz-López, J. L. Garrido, K. Benghazi, and L. Chung, "A Survey on Indoor Positioning Systems: Foreseeing a Quality Design", Distributed Computing and Artificial Intelligence, vol. 79, A. P. Leon F. de Carvalho, S. Rodríguez-González, J. F. Paz Santana, and J. M. C. Rodríguez, Eds. Berlin, Heidelberg: Springer Berlin Heidelberg, pp. 373-380, 2010.

[6]     C. Y. Chen, J. P. Yang, G. J. Tseng, Y. H. Wu, R. C. Hwang, and others, "An Indoor Positioning Techniques based on Fuzzy Logic," In the Proceeding(s) of the International Multi Conference of Engineers and Computer Scientists, vol. 2, 2010.

[7]     S. Woo et al., "Application Of Wifi-based Indoor Positioning System for Labor Tracking at Construction Sites: A Case Study in Guangzhou MTR", Automation in Construction, vol. 20, no. 1, pp. 3-13, 2011.

[8]     J. Yim, C. Park, J. Joo, and S. Jeong, "Extended Kalman Filter for Wireless LAN Based Indoor Positioning", Decision Support Systems, vol. 45, no. 4, pp. 960-971, 2008.

[9]     A. Mulloni, D. Wagner, I. Barakonyi, and D. Schmalstieg, "Indoor Positioning and Navigation with Camera Phones," IEEE Pervasive Computing, vol. 8, no. 2, pp. 22-31, 2009.

[10]   M. B. Kjærgaard, H. Blunck, T. Godsk, T. Toftkjær, D. L. Christensen, and K. Grønbæk, "Indoor Positioning Using GPS Revisited," Pervasive Computing, vol. 6030, P. Floréen, A. Krüger, and M. Spasojevic, Eds. Berlin, Heidelberg: Springer Berlin Heidelberg, pp. 38-56, 2010.

[11]   R. Hansen, R. Wind, C. S. Jensen, and B. Thomsen, "Seamless Indoor/Outdoor Positioning Handover for Location-Based Services in Streamspin," In the Proceeding(s) of the 10th International Conference on Mobile Data Management: Systems, Services and Middleware, pp. 267-272. 2009.

[12]   Hui Liu, H. Darabi, P. Banerjee, and Jing Liu, "Survey of Wireless Indoor Positioning Techniques and Systems," IEEE Transactions on Systems, Man, and Cybernetics, Part C: Applications and Reviews, vol. 37, no. 6, pp. 1067-1080, 2007.

[13]   W. M. Y. Wan Bejuri, M. M. Mohamad, and M. Sapri, "Ubiquitous Positioning: A Taxonomy for Location Determination on Mobile Navigation System," Signal & Image Processing: An International Journal, vol. 2, no. 1, pp. 24-34, 2011.

[14]   S. A. Shafer, "Using Color to Separate Reflection Components", Color Research & Application, vol. 10, no. 4, pp. 210–218, 1985.

[15]   G. J. Klinker, S. A. Shafer, and T. Kanade, "A Physical Approach to Color Image Understanding", International Journal of Computer Vision, vol. 4, no. 1, pp. 7–38, 1990.

[16]   J. Von Kries, Influence of Adaptation on The Effects Produced by Luminous Stimuli. Source of Color Vision (MIT Press), USA, 1970.

[17]   B. Funt and H. Jiang, "Nondiagonal Color Correction", In the Proceeding(s) of the International Conference on Image Processing, vol. 1, pp. 481-484.  2003

[18]   H. Y. Chong, S. J. Gortler, and T. Zickler, "The Von Kries Hypothesis and a Basis for Color constancy", In the Proceeding(s) of the 11th IEEE International Conference on Computer Vision, pp. 1–8, 2007.

[19]   G. D. Finlayson, M. S. Drew, and B. V. Funt, "Spectral Sharpening: Sensor Transformations for Improved Color Constancy", JOSA A, vol. 11, no. 5, pp. 1553–1563, 1994.

[20]   K. Barnard, F. Ciurea, and B. Funt, "Sensor Sharpening for Computational Color Constancy", Journal of the Optical Society of America A, vol. 18, no. 11, pp. 2728-2743, Nov. 2001.

[21]   G. West and M. H. Brill, "Necessary and Sufficient Conditions for Von Kries Chromatic Adaptation to Give Color Constancy", Journal of Mathematical Biology, vol. 15, no. 2, pp. 249–258, 1982.

[22]   J. A. Worthey and M. H. Brill, "Heuristic Analysis of Von Kries Color Constancy", JOSA A, vol. 3, no. 10, pp. 1708–1712, 1986.

[23]   G. D. Finlayson, M. S. Drew, and B. V. Funt, "Color Constancy: Generalized Diagonal Transforms Suffice", JOSA A, vol. 11, no. 11, pp. 3011–3019, 1994.

[24]   B. V. Funt and B. C. Lewis, "Diagonal Versus Affine Transformations for Color Correction", JOSA A, vol. 17, no. 11, pp. 2108–2112, 2000.

[25]   M. H. Brill, "Minimal Von Kries Illuminant Invariance", Color Research & Application, vol. 33, no. 4, pp. 320–323, 2008.

[26]   G. Buchsbaum, "A Spatial Processor Model for Object Colour Perception", Journal of the Franklin Institute, vol. 310, no. 1, pp. 1–26, 1980.

[27]   E. H. Land, The Retinex Theory of Color Vision, vol. 237. WH Freeman Company (Scientific American), USA, 1977.

[28]   A. Gijsenij and T. Gevers, "Color Constancy by Local Averaging", In the Proceeding(s) of the 14th International Conference on Image Analysis and Processing Workshops, pp. 171-174, 2007.

[29]   M. Ebner, "Color Constancy based on Local Space Average Color", Machine Vision and Applications, vol. 20, pp. 283-301, 2008.

[30]     B. Funt and L. Shi, "The Rehabilitation of maxRGB", In the Proceeding(s) of the IS&T/SID's Color Imaging Conference, 2010.

[31]     B. Funt and L. Shi, "The Effect of Exposure on maxRGB Color Constancy", In the Proceeding(s) of the  SPIE-IS&T Electronic Imaging, Vol. 7527, 2010.

[32]     G. D. Finlayson and E. Trezzi, "Shades of Gray and Colour Constancy", In the Proceeding(s) of the IS&T/SID Twelfth Color Imaging Conference, pp. 37–41, 2004.

[33]     J. van de Weijer, T. Gevers, and A. Gijsenij, "Edge-Based Color Constancy", IEEE Transactions on Image Processing, vol. 16, no. 9, pp. 2207-2214, 2007.

[34]     H. H. Chen, C.-H. Shen, and P.-S. Tsai, "Edge-based Automatic White Balancing with Linear Illuminant Constraint," In the Proceeding(s) of the SPIE, vol. 6508, pp. 65081D-65081D-12. 2007.

[35]     A. Chakrabarti, K. Hirakawa, and T. Zickler, "Color Constancy Beyond Bags of Pixels", In the Proceeding(s) of the IEEE Conference on Computer Vision and Pattern Recognition, pp. 1-6, 2008.

[36]     A. Gijsenij, T. Gevers, and J. Van De Weijer, "Physics-based Edge Evaluation for Improved Color Constancy", In the Proceeding(s) of the IEEE Conference on Computer Vision and Pattern Recognition, pp. 581-588, 2009.

[37]     H.-C. Lee, "Method for Computing The Scene-Illuminant Chromaticity from Specular Highlights," Journal of the Optical Society of America A, vol. 3, no. 10, pp. 1694-1699, 1986.

[38]     S. Tominaga and B. A. Wandell, "Standard Surface-Reflectance Model and Illuminant Estimation", Journal of the Optical Society of America A, vol. 6, no. 4, pp. 576-584, 1989.

[39]     H. Glenn, "Estimating Spectral Reflectance using Highlights", Image and Vision Computing, vol. 9, no. 5, pp. 333-337, 1991.

[40]     R. T. Tan, K. Ikeuchi, and K. Nishino, "Color Constancy Through Inverse-Intensity Chromaticity Space," Digitally Archiving Cultural Objects, pp. 323-351, 2008.

[41]     G. D. Finlayson and G. Schaefer, "Solving for Colour Constancy using a Constrained Dichromatic Reflection Model" , International Journal of Computer Vision, vol. 42, no. 3, pp. 127–144, 2001.

[42]     J. Toro and B. Funt, "A Multilinear Constraint on Dichromatic Planes for Illumination Estimation," IEEE Transactions on Image Processing, vol. 16, no. 1, pp. 92-97, 2007.

[43]     T. Javier, "Dichromatic Illumination Estimation without Pre-Segmentation", Pattern Recognition Letters, vol. 29, no. 7, pp. 871-877, 2008.

[44]     L. Shi and B. Funt, "Dichromatic Illumination Estimation via Hough Transforms in 3D", In the Proceeding(s) of the IS&T's European Conference on Color in Graphics, Imaging and Vision, 2008.

[45]     D. Comaniciu and P. Meer, "Mean Shift Analysis and Applications", In the Proceeding(s) of the Seventh IEEE International Conference on Computer Vision, vol. 2, pp. 1197–1203, 1999.

[46]     D. Guru, R. Dinesh, and P. Nagabhushan, "Boundary Based Corner Detection and Localization using New Cornerity Index: A Robust Approach", In Proceeding(s) of the First Canadian Conference on Computer and Robot Vision, pp. 417–423, 2004.

[47]     P. Bahl and V. N. Padmanabhan, "RADAR: an in-Building RF-based User Location and Tracking System", In Proceeding(s) of the IEEE INFOCOM 2000 Nineteenth Annual Joint Conference of the IEEE Computer and Communications Societies, vol. 2, pp. 775-784 vol.2, 2000.

[48]     M. A. Fischler and R. C. Bolles, "Random Sample Consensus: a Paradigm for Model Fitting with Applications to Image Analysis and Automated Cartography", Communications of the ACM, vol. 24, no. 6, pp. 381–395, 1981.

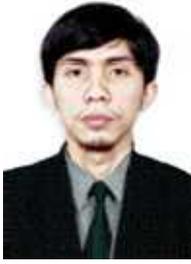

**Wan Mohd Yaakob Wan Bejuri** received his Bachelor Degree in Computer Science from Universiti Teknologi Malaysia in 2009. Previously, he was received Diploma in Electronic Engineering in 2005 from Politeknik Kuching Sarawak. He is currently Master Candidate at Universiti Teknologi Malaysia and associate member of IEEE. For detail, please visit to: http://teknologimalaysia.academia.edu/WANMOHDYAAKOBWANBEJURI or email to wanmohdyaakob@gmail.com or wanmohdyaakob@ieee.org

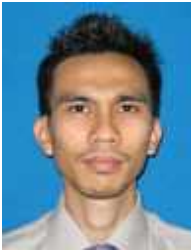

**Mohd Murtadha Mohamad** received his Ph.D. in Electrical and Master Degree in Embedded System Engineering from Heriot-Watt University. He received his Bachelor Degree in Computer from Universiti Teknologi Malaysia. He is currently a senior lecturer at Universiti Teknologi Malaysia and also a member of IEEE. For detail, please visit to: http://csc.fsksm.utm.my/murtadha/

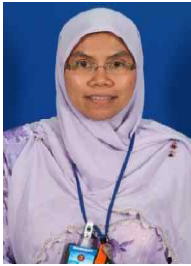

**Maimunah Sapri** received her Ph.D. in Real Estate Management from Heriot-Watt University. She received her Master Degree in Facilities Management and Bachelor Degree of Surveying in Property Management from Universiti Teknologi Malaysia. Previously, she received Diploma in Estate Management from Institute of Technology MARA. She is currently a senior lecturer at Universiti Teknologi Malaysia. For detail, please visit to: http://www.cresutm.com/

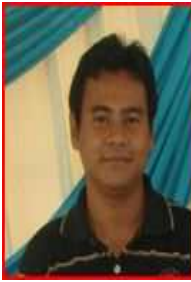

**Mohd Adly Rosly** received the Bachelor Degree of Engineering in Geomatics from Universiti Teknologi Malaysia in 2011. He is currently Research Assistant Officer at Universiti Teknologi Malaysia. For detail please visit: http://teknologimalaysia.academia.edu/MohdAdlyRosly